\documentclass[letterpaper, 10 pt, conference]{ieeeconf}  

\IEEEoverridecommandlockouts                              

\usepackage{graphics} 
\usepackage{graphicx}

\usepackage{epsfig} 
\usepackage{amsmath} 
\usepackage{amssymb}  
 \usepackage{algorithm}
\usepackage[compatible]{algpseudocode}
\algnewcommand\AAND{\textbf{ and }}
\algnewcommand\Or{\textbf{ or }}
\usepackage{color}
\usepackage{citesort}
\usepackage{flushend}
\usepackage{url}
\usepackage[table,xcdraw]{xcolor}
\usepackage{amsmath}

\usepackage[colorinlistoftodos]{todonotes}

\DeclareMathAlphabet{\pazocal}{OMS}{zplm}{m}{n}

\DeclareMathAlphabet{\mathpzc}{OT1}{pzc}{m}{it}

\usepackage{array}
\newcolumntype{C}[1]{>{\centering\arraybackslash}p{#1}}
\newcolumntype{M}[1]{>{\raggedright\arraybackslash}p{#1}}

\usepackage{array} 
\newcolumntype{L}[1]{>{\raggedright\let\newline\\\arraybackslash\hspace{0pt}}m{#1}}	
\newcolumntype{S}[1]{>{\centering\let\newline\\\arraybackslash\hspace{0pt}}m{#1}}
\newcolumntype{R}[1]{>{\raggedleft\let\newline\\\arraybackslash\hspace{0pt}}m{#1}}

\usepackage[nolist,nohyperlinks]{acronym}
\acrodef{mlt}[MLTs]{Martian Lava Tubes}

\makeatletter
\renewcommand*{\@opargbegintheorem}[3]{\trivlist
  \item[\hskip \labelsep{\itshape #1\ #2}] \textit{(#3)}\ }
\makeatother

\title{\LARGE \bf
 Design and Experimental Verification of a Jumping Legged Robot for Martian Lava Tube Exploration
}

\title{\LARGE \bf
 Design and Experimental Verification of a Jumping Legged Robot for Martian Lava Tube Exploration
}

\author{J{\o}rgen Anker Olsen and  Kostas Alexis
\thanks{
}%
\thanks{The authors are with the Autonomous Robots Lab, NTNU, O.S. Bragstads Plass 2D, 7034, Trondheim, NO {\tt\small jorgen.a.olsen@ntnu.no} \newline
}
}

\begin{document}

\maketitle
\thispagestyle{empty}
\pagestyle{empty}

\begin{abstract}

The potential of Martian lava tubes for resource extraction and habitat sheltering highlights the need for robots capable to undertake the grueling task of their exploration. Driven by this motivation, in this work we introduce a legged robot system optimized for jumping in the low gravity of Mars, designed with leg configurations adaptable to both bipedal and quadrupedal systems. This design utilizes torque-controlled actuators coupled with springs for high-power jumping, robust locomotion, and an energy-efficient resting pose. Key design features include a 5-bar mechanism as leg concept, combined with springs connected by a high-strength cord. The selected 5-bar link lengths and spring stiffness were optimized for maximizing the jump height in Martian gravity and realized as a robot leg. Two such legs combined with a compact body allowed jump testing of a bipedal prototype. The robot is $0.472 \ \textrm{m}$ tall and weighs $7.9\ \textrm{kg}$. Jump testing with significant safety margins resulted in a measured jump height of $1.141 \ \textrm{m}$ in Earth's gravity, while a total of $4$ jumping experiments are presented. Simulations utilizing the full motor torque and kinematic limits of the design resulted in a maximum possible jump height of $1.52 \ \textrm{m}$ in Earth's gravity and $3.63 \ \textrm{m}$ in Mars' gravity, highlighting the versatility of jumping as a form of locomotion and overcoming obstacles in lower gravity.

\end{abstract}

\section{INTRODUCTION}\label{sec:intro}

Robotic systems have long been utilized as means to access remote and strenuous environments including for planetary exploration. The recent experience of the DARPA Subterranean Challenge~\cite{tranzatto_cerberus_2022} demonstrated the potential of autonomous robots, and in particular legged systems, to explore complex underground settings such as natural cave networks~\cite{dang_graph-based_2020}. Motivated by this success, roboticists are exploring the use of legged robots to explore the scientifically promising \ac{mlt}~\cite{agha2021nebula,whittaker_technologies_2012}. \ac{mlt} are volcanic cavern formations recognized based on images from the Viking orbiter and subsequent orbiter missions~\cite{sauro_lava_2020}. Their exploration is important as they can open new avenues for the scientific study of the geological, paleohydrological, and supposed biological history of the planet, while also potentially offering resources and a ``safe harbor'' to host a possible human settlement protected from UV radiation, dust and other high-risk phenomena on Mars' surface~\cite{leveille_lava_2010}.

\begin{figure}[!ht]
\centering
    \includegraphics[width=0.99\columnwidth]{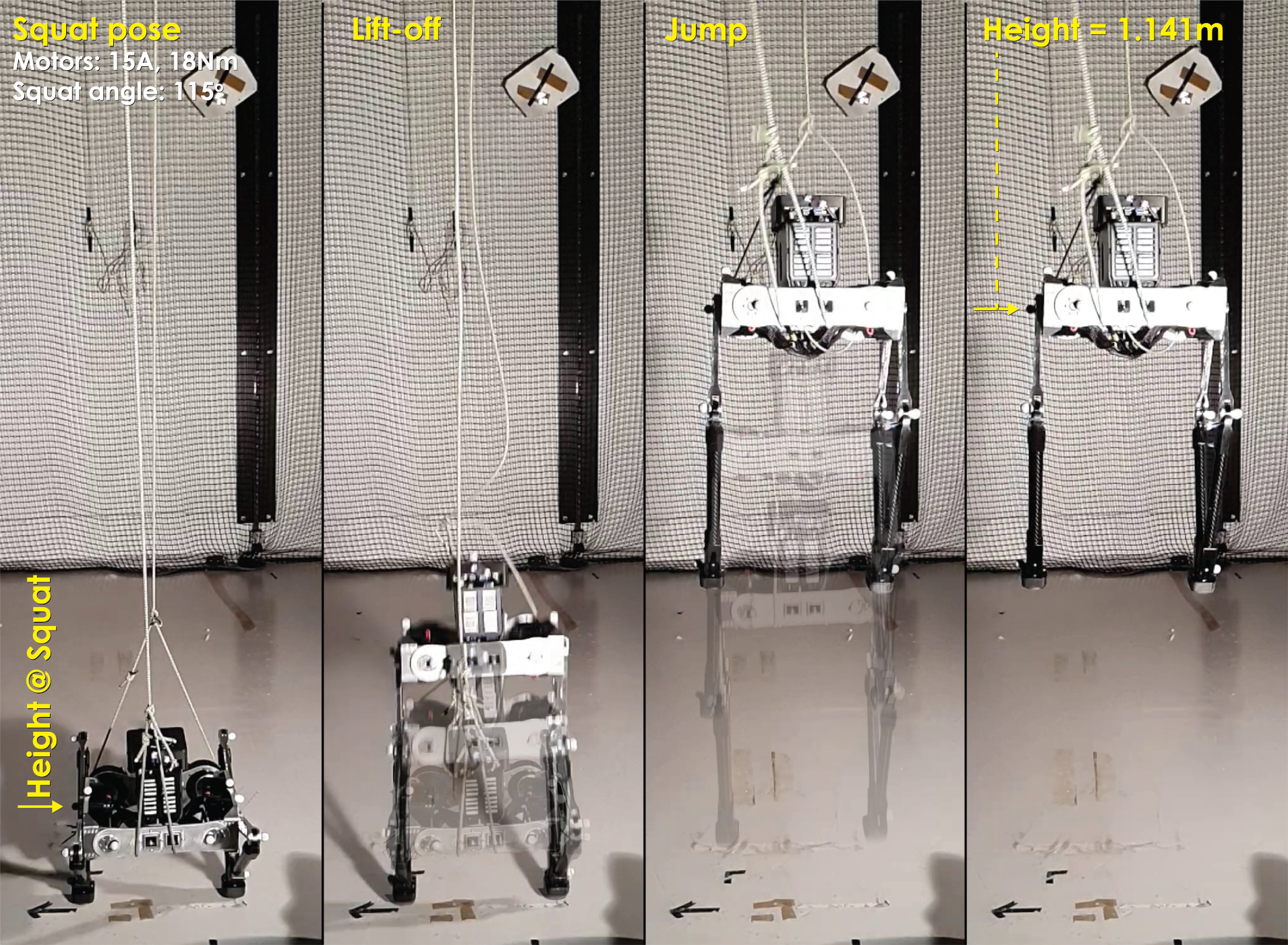}
\vspace{-5ex} 
\caption{Powered jump reaching $1.141\ \textrm{m}$ high with the motors' torque at $18\ \textrm{Nm}$ (max: $24.8\ \textrm{Nm}$) and the squat angle being $115\ \textrm{deg}$ (max: $120\ \textrm{deg}$). The maximum jumping capacity in Earth's gravity is derived to be $1.52\ \textrm{m}$, while on Mars this is increased to $3.63\ \textrm{m}$ for the complete bipedal system.}
\label{fig:OLYMPUSJUMP_Frpnt}
\vspace{-5ex} 
\end{figure}

Space exploration in the modern age is dominated by the use of robotic systems, primarily rovers, but recently also helicopters roaming to explore the surface of other planets~\cite{balaram_mars_2018}. However, the exploration of \ac{mlt} represents a different, grueling challenge. First, the robots have to be able to explore large-scale underground geometries, which in turn is likely to render flying systems unable to undertake the mission fully on their own. Second, surface access to \ac{mlt} is only possible through confined passages such as skylights. Third, \ac{mlt} are expected to involve much larger cross sections (compared to lava tubes on Earth) due to the reduced gravity of Mars ($38\%$ that of Earth) and uneven terrain geometries~\cite{sauro_lava_2020}. The latter two features of \ac{mlt} entail the need for small robots that, however, can a) present the endurance necessary for long-term exploration, and b) the ability to conquer the challenging terrain expected to involve large-scale obstacles and complicated geometries. This is likely to limit the use of traditional rovers or even conventional legged robots tailored to walking-only locomotion.

Motivated by the above, this work contributes the design of a new legged system optimized for jumping-based locomotion tailored to the low-gravity, massive scale, and complex terrain geometry expected in \ac{mlt}. Employing a 5-bar parallel closed kinematic chain for each leg, the design parameters are chosen such that vertical jumping capacity is maximized, exploiting Mars' low gravity, while at the same time dexterous walking locomotion remains possible~\cite{kenneally_design_2016}. Jumping-based locomotion could prove critical both as means to overcome very large obstacles and also as an efficient strategy for long foraging motions in the low-gravity environment of \ac{mlt}. Optimized in simulation and verified in experimental studies, the system can jump up to $1.52\ \textrm{m}$ high on Earth, up to $3.63\ \textrm{m}$ on Mars gravity, and was experimentally demonstrated with a jump of $1.141\ \textrm{m}$ (subject to Earth's gravity) without reaching the maximum of its actuators' capacity or the limits of its kinematics.

The remainder of this paper is organized as follows: Section~\ref{sec:related} presents related work, followed by the system design in Section~\ref{sec:sys_des}. Simulation-based evaluations are presented in Section~\ref{sec:sim}, while experimental results are detailed in Section~\ref{sec:experiments}. Finally, conclusions are drawn in Section~\ref{sec:conclusions}.

\section{RELATED WORK}\label{sec:related}

The modern era of space exploration leverages robotic systems, such as rovers and helicopters, to probe extraterrestrial terrains such as on Mars. Despite years of research, a host of challenges persist. Of particular interest to this work is the capacity of robotic systems to access and explore off-world lava tubes and especially those of Mars. Due to narrow skylight access, robots tasked to enter would have to be small in size rendering traditional designs such as conventional rovers inapplicable as in small scale they also present serious terrain traversability limitations~\cite{sauro_lava_2020}. Custom designs such as Axel/DuAxel~\cite{nesnas2012axel} provide solutions to specific issues like narrow skylight access but this robot eventually inherits the traversability limitations of rovers. Other custom concepts such as ReachBot~\cite{newdick2023designing} likewise may present advantages (e.g., traversability compared to rovers),  but their potential limitations are not well understood. Motivated by the above and the demonstrated performance of legged robots in conquering the challenges of the DARPA Subterranean Challenge --which among others focused on mapping natural cave networks-- researchers are now actively looking into the possibility of utilizing legged systems for \ac{mlt} exploration~\cite{agha2021nebula}.

In view of the above and given our current understanding of the vast geometries of \ac{mlt}, this work specifically relates to the class of legged robots with increased jumping capabilities~\cite{zhang2020biologically}. Among the most relevant systems, SpaceBok~\cite{arm_spacebok_2019}, with a design motivated by the need for high jumping on Moon gravity ($1.62\ \textrm{m/s}^2$, $17\%$ that of Earth)~\cite{kolvenbach2019towards,rudin2021cat} employs a 4-bar design and a stabilizing reaction wheel. At its maximum jumping tests, its center of mass reaches at approximately $1.05\ \textrm{m}$. Minitaur~\cite{kenneally_design_2016,kenneally_leg_2015} with a diamond leg demonstrated a jump of $0.48\ \textrm{m}$. Aiming to assess the agility of legged robots, the authors in~\cite{caluwaerts2023barkour} introduced the ``Barkour'' benchmark which includes the ``broad jump'' where the robot has to clear a $0.5$ m long and $1$ m wide board. Employing a conventional leg design, the work in~\cite{ding2017design} realized a small and lightweight leg evaluated to jump up to $0.62\ \textrm{m}$. This work further presents a methodology to optimize the jumping maneuver. Likewise, the authors in~\cite{nguyen2022contact} present contact-timing optimization enabling a Unitree A1 legged robot to make jumps of limited height. Earlier, the method in~\cite{nguyen2019optimized} enabled a Mini Cheetah robot to show leaping capabilities and jump onto a 30 inch ($\approx 0.76\ \textrm{m}$) table. Also using a conventional leg, but combined with wheels, Ascento~\cite{klemm2019ascento} utilizes small jumps to deal with obstacles such as stair steps. The works in~\cite{ye2021modeling,shen2020optimized} further contribute into methods for optimizing the jumping maneuver but without consideration on the mechatronic system capable for high jumping. Focusing on bipedal humanoids, the work in~\cite{qi2023vertical} demonstrated $0.5\ \textrm{m}$ vertical jump, alongside simulations of forward jumping. Relevant is also the experience from results of the Atlas humanoid albeit with hydraulic actuation~\cite{atlas_website}. Hydraulic actuation prevailed in the early years of legged robot research~\cite{raibert_legged_2000} but currently high-torque electric actuators are more common in legged robots such as ANYmal~\cite{hutter_anymal_2016}, Boston Dynamics Spot~\cite{spot_website}, and Unitree A1~\cite{unitree_a1_website}. On the miniature-scale, a host of works present systems that weigh less than $100\ \textrm{g}$ and demonstrate remarkable jumping ($1.25\ \textrm{m}$ in~\cite{haldane_repetitive_2017}, $1.62\ \textrm{m}$ in~\cite{jung2016integrated} and $3.1\ \textrm{m}$ for the locust-inspired system in~\cite{zaitsev2015locust}) given the relative power density of actuation for the weight~\cite{kovac2008miniature,haldane_repetitive_2017,zhao_miniature_2014,jung2016integrated,zaitsev2015locust,xu2023design,zhao2013msu}.  Naturally, such systems are limited regarding their sensing and overall payload, as well as their endurance. This work focuses on a design with high jumping capacity combined with retaining the walking dexterity that the 5-bar diamond leg offers~\cite{zhao_miniature_2014}, as well as significant payload capabilities, alongside simplicity and a low-cost as it employs off-the-shelf electric drives. The aforementioned jump tests were performed in Earth gravity.

\section{SYSTEM DESIGN}\label{sec:sys_des}
This section outlines the robot design and leg optimization. 

\subsection{Design goals, constraints, and operational assumptions}

The robotic system is purposefully engineered with the principal objective of achieving high jumping capacity both on Earth and especially in the low-gravity Mars environment, enabling the traversal of obstacles taller than itself. A successful realization of this goal necessitates the following functionalities: First, the robot should be capable of significant and consequential vertical and forward jumps. Second, post-departure from the ground, the leg articulation should enable the robot to manipulate leg movements to counteract any initial rotation induced by the jumping motion~\cite{rudin2021cat}. This functionality is vital to ensure that the robot lands in a desirable configuration (i.e., legs toward the ground) and avoids crash landings. Third, the design should incorporate strategies to recover energy post-jump by being designed with high compliance to tolerate collisions, meaning the leg architecture should leverage integrated springs and active motor damping for energy recovery during landing to enable the possibility of consecutive jumps.

Further, the design aims to furnish the robot with standard walking capabilities and a low-energy consumption resting posture. To actualize these design goals, key components have been integrated, namely a) high-torque actuators for leg actuation facilitating both jumping and walking, and b) an aiding spring system designed to augment the robot's jumping capabilities while adding compliance for landing. Moreover, this non-actuated force applied to the robot leg as it stands contributes to energy conservation by reducing the demand on the motors while standing statically to almost zero, utilizing the angle of the force on the knee joints and motors, resulting in two stable spring configurations, standing upright and a more stable squatting pose.

The design of robot legs varies among different legged systems. MIT mini cheetah~\cite{katz_mini_2019}, Boston Dynamics Spot~\cite{spot_website}, and ANYbotics ANYmal~\cite{hutter_anymal_2016} use serial linkage with actuators in the hip and/or knee. On the other hand, SpaceBok~\cite{arm_spacebok_2019} and Minitaur~\cite{kenneally_design_2016} use a closed kinematic chain. SpaceBok, also designed for jumping maneuvers, employs a 4-bar design. For the contributed leg design, a 5-bar parallel motion mechanism was chosen for its advantages in the robot's workspace, as well as the increased range of motion, allowing for more effective motor usage during a jump~\cite{kenneally_design_2016,kenneally_leg_2015,dong2017design}. Motivated by the results in~\cite{kenneally_design_2016,kenneally_leg_2015} a diamond 5-bar configuration is selected as it also offers good walking performance. The 5-bar design is depicted in Figure \ref{fig:UnloadedKinematicChain}.

The robot uses three actuators per leg to enable three actuated degrees of freedom. One actuator is for hip abduction, while the other two are utilized in the 5-bar parallel mechanism for walking and jumping. All three are to be used while performing in-air stabilization and dynamic walking.

\subsection{Electrical system and actuation control}

As actuators, we selected the CubeMars AK70-10, a commercial off-the-shelf motor that is relatively low-cost and provides high torque, with a maximum torque output of $24.8\ \textrm{Nm}$. This actuator comes with a built-in encoder and is controlled over CAN-bus at a maximum rate of $1\ \textrm{MHz}$. The motor has a built-in planetary gear with a 10:1 reduction ratio. The robot employs a high-energy peak consumption strategy, leveraging the maximum torque capabilities of the actuators in short bursts when performing jumps. Each actuator can draw up to $20.5\ \textrm{A}$ to achieve peak torque at peak output, with an aggregate of $123\ \textrm{A}$ for all six motors (bipedal robot realization). To be able to deliver this and have a significant safety margin (in the case of $12$ motors for a quadruped), two Tattu R-Line Version 5.0 $1200 \ \textrm{mAh}$ $22.2\ \textrm{V}$ $150$C 6S1P Lipo Batteries were wired in series, constituting a $12$S, $44.4 \ \textrm{V}$ setup. Further parallel connection resulted in a capacity of $2400 \ \textrm{mAh}$ and a current delivery capacity of $ 150 \ \textrm{C}$. This configuration offers a substantial safety margin for a 6-motor bipedal robot and can accommodate a 12-motor system corresponding to a quadruped option. The electrical power from these batteries is distributed to the motors via a dedicated Power Distribution Block (PDB). The PDB also channels power to a DC-DC converter, which outputs $15 \ \textrm{V}$ to the onboard Asus Next Unit of Computing (NUC) with an AMD Ryzen 7 5700U CPU. This NUC, functioning as the core processing unit, runs the ROS-based motor jump controller commanding the motors and produces the required CAN-bus signal via an Innomaker USB-to-CAN converter providing the motor driver board with the necessary torque commands to perform desired movements. A VectorNav VN-100 is used as an onboard Inertial Measurement Unit (IMU) to capture relevant motion data. The C++ ROS-based controller runs at $500\ \textrm{Hz}$ on the NUC with a PID controller used to control the motors for workspace and jump testing.

\subsection{Kinematics of a single robot leg on a plane}
\begin{figure}[h!]
\vspace{-3ex} 
\centering
    \includegraphics[width=0.81\columnwidth]{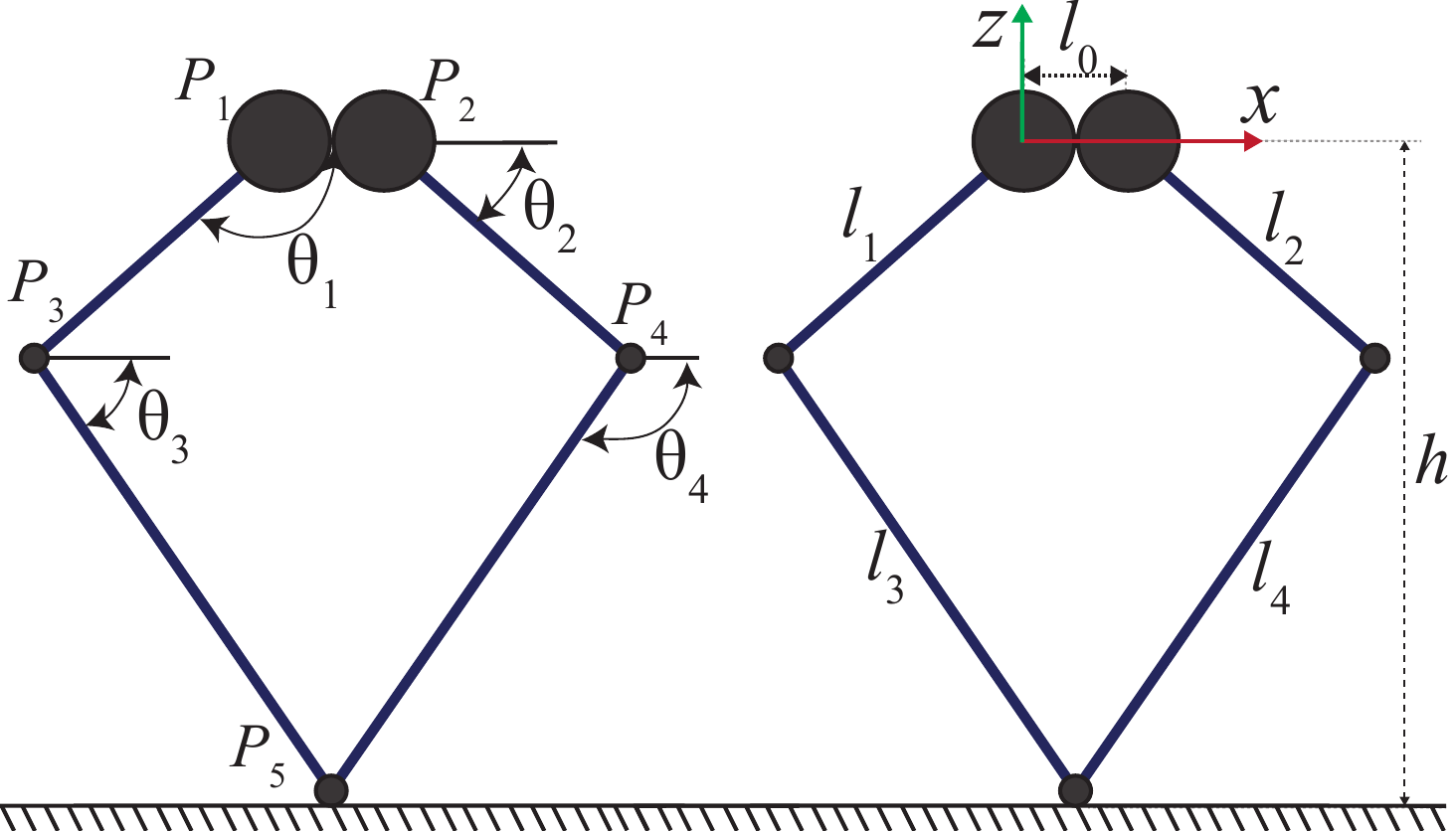}
\vspace{-2ex} 
\caption{Kinematic chain representing the 5-bar design. We set $l_1 = l_2$ and $l_3 = l_4$. Note that the $Y$-axis pointing out of the figure.}
\label{fig:UnloadedKinematicChain}
\vspace{-1ex} 
\end{figure}

The 5-bar mechanism concept for the robot leg is illustrated as a simplified 2D sketch in Figure \ref{fig:UnloadedKinematicChain}. The endpoint of the paw is treated as an end-effector, and its position ${[x_p \ z_p]}^T$ on the $XZ$-plane can be described as in~\cite{zi_dynamic_2011}:

\begin{equation}
\begin{split}
    \begin{bmatrix}
        x_p\\
        z_p
    \end{bmatrix}
         &= 
    \begin{bmatrix}
        l_1 \cos(\theta_1) + l_3 \cos(\theta_3) \\
        l_1 \sin(\theta_1) + l_3 \sin(\theta_3)
    \end{bmatrix}
     \\ &= 
    \begin{bmatrix}
        l_0 + l_2 \cos(\theta_2) + l_4 \cos(\theta_4)\\
        l_2 \sin(\theta_2) + l_4 \sin(\theta_4)
    \end{bmatrix}
\end{split}
\end{equation}
where $l_0$, $l_1$, $l_2$, $l_3$, and $l_4$ are the lengths of the 5-bar leg, and $\theta_1 $, $\theta_2$, $\theta_3$, and $\theta_4$ are the joint angles as seen in Figure~\ref{fig:UnloadedKinematicChain}. The forward kinematics can then be described as given only the motor angles  $\theta_1 $ and $\theta_2$ ($\theta_3 $ and $\theta_4$ are calculated) thus $\theta_3 $ and $\theta_4$ can be derived as follows:

\small
\begin{equation}
    \theta_4 = 2 \arctan ( \frac{a \pm \sqrt{a^{2} + b^{2} - c^{2}}}{b-c})
\end{equation}
\normalsize

\small
\begin{equation}
\begin{split}
         a &= 2 l_4 l_2 \sin (\theta_2) - 2 l_1 l_4 \sin (\theta_1) \\
         b &= 2 l_4 l_0 - 2 l_1 l_4 \cos(\theta_1) + 2 l_4 l_2 \cos(\theta_2)\\
         c &= l_0^{2} + l_1^{2} + l_2^{2} - l_3^{2} + l_4^{2} - 2l_1 l_2 \sin(\theta_1) \sin(\theta_2)\\
           &- 2l_1l_0 \cos(\theta_1) + 2l_2 l_0 \cos(\theta_2) - 2 l_1 l_2 \cos(\theta_1) \cos(\theta_2)
\end{split}
\end{equation}
\normalsize

Then $\theta_3$ is given by the following expression:
 
\small
\begin{equation}
     \theta_3 = \arcsin (\frac{l_4 \sin(\theta_4) + l_2 \sin(\theta_2) - l_1\sin(\theta_1)}{l_3})
\end{equation}
\normalsize

\normalsize

To determine the link lengths needed to achieve the maximum jump height given the 5-bar parallel closed kinematic chain and the incorporation of spring support, a grid-search optimization was performed to determine the optimal link lengths, and link ratios. It is set that $l_1 = l_2$ and $l_3 = l_4$ as a diamond 5-bar configuration is employed. The spring stiffness was further co-optimized as subsequently explained. 

\subsection{Leg Parameters Optimization in Simulation}

In order to determine optimized link lengths and spring stiffness both for a single leg and a bipedal robot model, an extensive simulation study was performed in MATLAB Simscape considering a bipedal robot with two identical legs and the hip joints locked during tests. During these simulations, the robot's body and the end-effector were constrained to one-dimensional vertical movement. The grid search optimization simulations were performed in both Earth gravity and Mars gravity. The simulations were performed with a maximum torque saturation of $22.5\ \textrm{Nm}$, $9$ \% less than the max output of the AK70-10 V2.1 to offer some safety margin. The results from the simulation (through grid search) were used to determine the link lengths $l_1$, $l_2$, $l_3$, $l_4$, and spring stiffness $k$, resulting in an optimal link length ratio for vertical jumping capacity for the 5-bar parallel design. The link $l_0$ is set to $l_0 = 0.09 \  \textrm{m} $ throughout this study. 

Physical system modeling was used to build up a model of the 5-bar design in Simscape where the body, hips, calves, and paws were connected in a serial link as necessitated by the 5-bar design. In Simulink, a variable step solver, switching between \texttt{ode23} and \texttt{ode45}, was used to optimize accuracy and performance when switching hybrid dynamics occurred between the states of ground contact during stands, liftoff, no contact while in air, and then ground contact when landing. Some virtual ground penetration was allowed to simulate the compression of the compliant part of a paw. 

Simulations were performed with values for the ball bearing friction and the motor damping coefficients being acquired through preliminary experiments on the physical leg and bipedal prototype described in Section~\ref{sec:bipedalrealization}. The range of motion of each leg was constrained to what the real system would be able to achieve. Essentially, these values were tuned to fit with the experimental results.

\begin{figure}[h!]
\vspace{-3ex} 
    \centering
        \includegraphics[width=0.99\columnwidth]{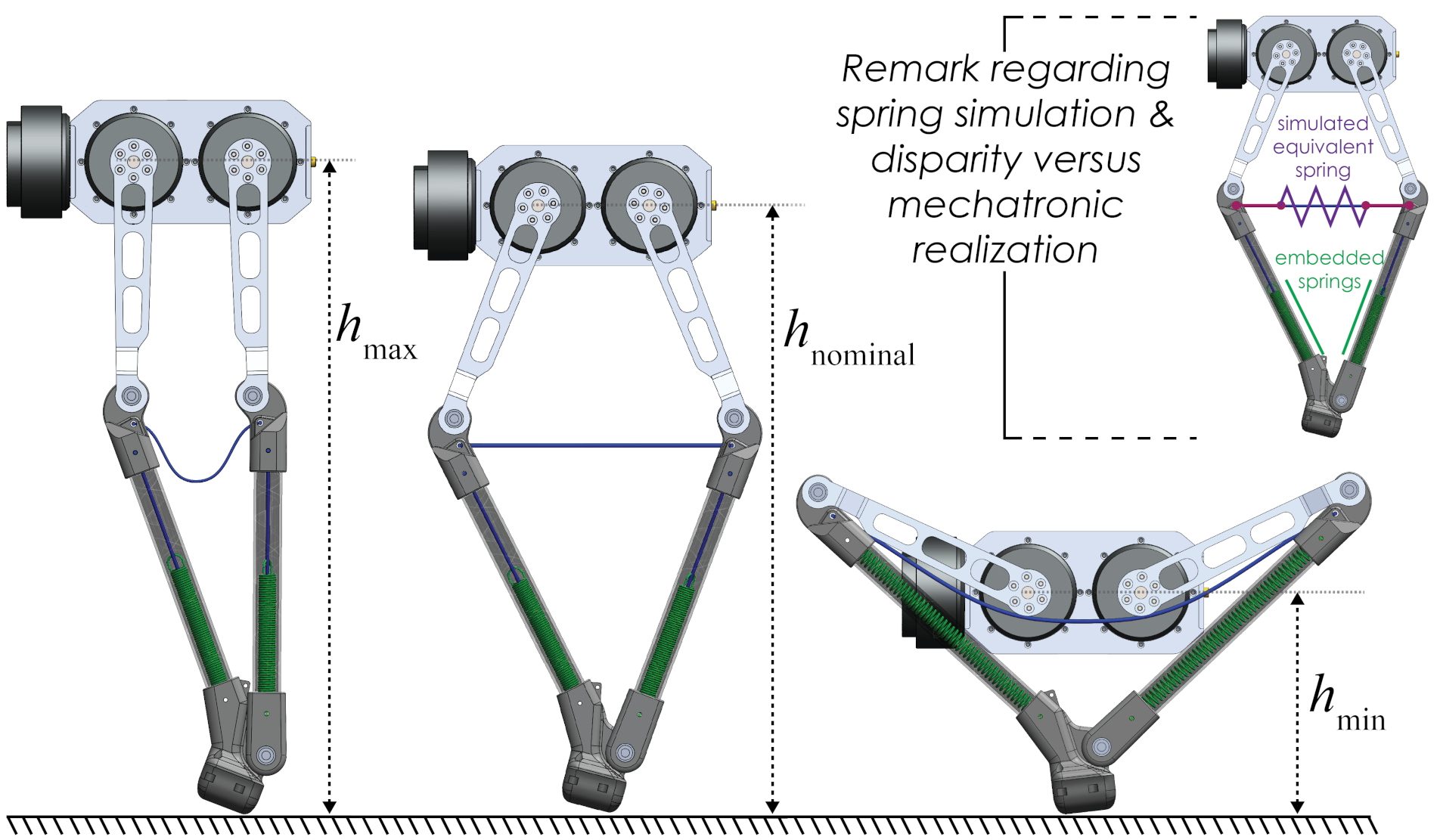}
    \vspace{-4ex} 
    \caption{Robot leg configurations representing leg configurations for simulation optimization and jump tests. For the leg realization, the spring is located inside the carbon tube calf at the bottom near the paw. In the two left configurations the spring is not significantly extended. In the squat pose it is fully extended. The springs can be seen in the semi-translucent carbon-tube calf. Top right: it is remarked that although $2$ springs connected with a paracord string are used on the real robot, in simulation an approximately-equivalent spring connecting the two knee joints is used instead. }
    \label{fig:SimPoses}
    \vspace{-4ex} 
\end{figure}

The spring mechanism was simulated as one single spring connected between the knee joints of the 5-bar design. The real robot has two springs per leg located in each calf connected with a paracord string, as seen in Figure~\ref{fig:SimPoses}. Thus, it can be simplified to two springs in series (or an equivalent spring as visualized in Figure~\ref{fig:SimPoses}), indicating that the physical robot will need to have a spring with twice the spring stiffness of the simulation. The mechatronic realization with two embedded springs was opted for as it allows for a higher range of squat angles and improved mechanical reliability.

Figure \ref{fig:SimPoses} illustrates the starting point of the simulation. The legs are both pointing directly downward. For the grid search, this defines the zero position. Then, the PID controller is given setpoints of $17.5 \ \textrm{deg}$ where spring engagement starts as the spring natural length is set to $200\  \textrm{mm}$, resulting in $h_{\textrm{nominal}}$, subsequently, a squatting movement is executed over a span of $1.3 \ \textrm{sec}$ until the system reaches $\theta_1 = -120 \  \textrm{deg} $ and $\theta_2 = 120 \ \textrm{deg}$ for both legs. This squat posture is sustained for an additional $0.5\ \textrm{sec}$ prior to issuing the jump command. Upon receipt of the jump command, the setpoint reverts to $17.5\  \textrm{deg}$ as the motors actuate to their specified torque output saturation. Certain combinations of link lengths and spring stiffness may lead to scenarios where either the required force for full spring extension exceeds the motor capabilities or the squat results in a knee joint angle that inhibits the motors' ability to revert to a standing position. 

Note that the spring's natural length was selected primarily on the basis of a) the desired overall leg size (approximately similar to those of robots such as ANYmal or Spot), and b) the physical leg utilizing springs connected with a cord with some slack, as seen in Figure \ref{fig:SimPoses}, lending the leg the capability to have some basic range of motion to perform simple walking without engaging the springs. Other factors were the availability of off-the-shelf components and the torque limits of the selected motor. 

Table \ref{tab:OLYMPUSsimulationparameters} lists the basic parameters for the Simscape leg model and the ranges searched to find the optimal parameter values for the optimization of the legs' design.

\begin{table}[ht]
\begin{center}
    
\caption{Robot leg simulation parameters}
\label{tab:OLYMPUSsimulationparameters}
\begin{tabular}{l|c|c}
Parameter          & Symbol & Value             \\ \hline
Mass single leg    & $m_l$   & $3.3$ {[}kg{]}       \\
Mass of electronics and battery & $m_e$ & $1.3$ {[}kg{]} \\
Mass bipedal robot & $m_b$  & $7.9$ {[}kg{]}       \\
Link 0 length      & $l_0$   & $0.09$ \ {[}m{]}  \\
Link 1 length      & $l_1$   & $0.10-0.30$ \ {[}m{]}  \\
Link 2 length      & $l_2$   & $0.10-0.30$ \ {[}m{]}  \\
Link 3 length      & $l_3$   & $0.15-0.45$ \ {[}m{]}  \\
Link 4 length      & $l_4$   & $0.15-0.45$ \ {[}m{]}  \\
Spring Stiffness   & $k$     & $600-1000$ {[}N/m{]} \\
\end{tabular}
\end{center}
\end{table}

\begin{figure*}[h]
    \centering
    \includegraphics[width = 0.98\textwidth] {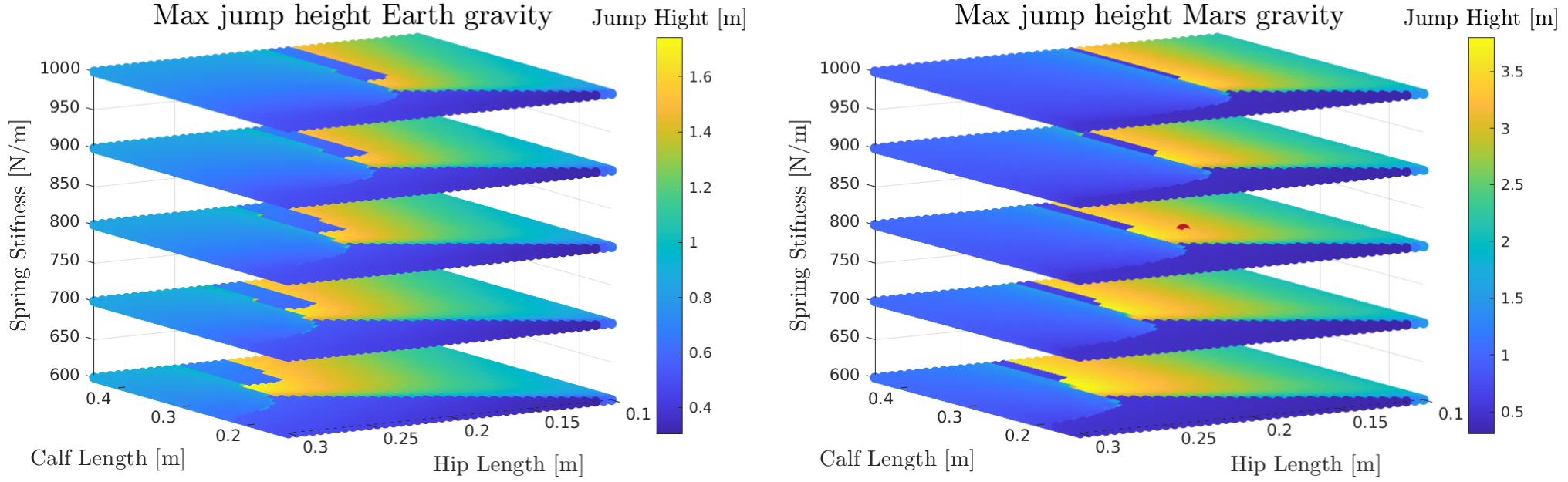}
        \vspace{-2ex} 
    \caption{4D grid search optimization with varying link lengths for hip and calf, and varying spring stiffness for the bipedal system. To the left, simulations are performed with Earth gravity (9.81 $\frac{m}{s^{2}}$), to the right with Mars gravity (3.71 $\frac{m}{s^{2}}$). The chosen link length combination is indicated in the Mars gravity simulations to the right with values in Table \ref{tab:OLYMPUSOptimizedValues}. It is noted that although few designs presented slightly higher jumping performance, it was opted to select link lengths that not only allowed high performance but also robust performance across a range of spring stiffness values and allowed the robot to squat without having to output torque close to the max torque limit, enabling more control authority in instances of unequal torque requirements for the motors.}
    \label{fig:4DOptimisation}
    \vspace{-3ex}
\end{figure*}

The result measured from each simulation iteration in the grid search optimization is the achieved height of the body (specifically of Motor 2 as in Figure~\ref{fig:SimPoses}) and the ground clearance measured from the maximum paw height. The configuration for the leg in the air until it lands, is the same as $h_{\textrm{nominal}}$. Figure~\ref{fig:4DOptimisation} depicts the resulting maximum jump heights measured from the body to the ground in a $4$D plot where each point is the maximum jump height achieved for each simulation, and its intensity is the jump height.

From the analysis of the simulation data, the optimized values for the link lengths and the spring stiffness were derived as in Table~\ref{tab:OLYMPUSOptimizedValues}. The selected parameters also accounted for performance robustness as outlined in Figure~\ref{fig:4DOptimisation}. The robot realization is based on these parameters. Simulations also indicated that a link length ratio ($l_1/l_3$) of $\sim $0.63 resulted in the optimized jumps for this symmetric 5-bar configuration. The numerical results for the presented simulations, alongside comparisons with optimizing a non-symmetric design similar to~\cite{arm_spacebok_2019} which achieved inferior jumping performance for similar leg size and actuation are available at \url{https://github.com/ntnu-arl/jumping-robots}.

\begin{table}[h]
\centering
\caption{Chosen optimized leg design values}
\vspace{-2ex}
\label{tab:OLYMPUSOptimizedValues}
\begin{tabular}{l|c|c}
Parameter          & Symbol & Value             \\ \hline
Link length ratio  & $\frac{l_1}{l_3}$  &  $\sim0.63 $ \\
Link 0 length      & $l_0$   & 0.09 {[}m{]}  \\
Link 1 length      & $l_1$   & 0.18 {[}m{]}  \\
Link 2 length      & $l_2$   & 0.18 {[}m{]}  \\
Link 3 length      & $l_3$   & 0.30 {[}m{]}  \\
Link 4 length      & $l_4$   & 0.30 {[}m{]}  \\
Spring Stiffness   & $k$     & 800 {[}N/m{]} \\
\end{tabular}
\vspace{-4ex}
\end{table}

\subsection{Bipedal Robot Realization for Vertical Jumping}\label{sec:bipedalrealization}

Taking into consideration the aforementioned design specifications for the 5-bar robot leg design, link lengths, spring stiffnesses, and minimum of 3 degrees of movement per leg, a bipedal robot was realized in practice. A disparity from the values in Table~\ref{tab:OLYMPUSOptimizedValues} is that for the real robot the selected spring stiffness is $870\ \textrm{N/m}$ for each leg leading to an equivalent value of $435\ \textrm{N/m} $ as per the simulation collective spring modeling. This decision was driven by the need to achieve a more damped and thus safer experimental jumping response. Furthermore, it is pointed out that the main force responsible for vertical leaps comes from the motor actuation.

\begin{figure}[h!]
    \centering
        \includegraphics[width=0.86\columnwidth]{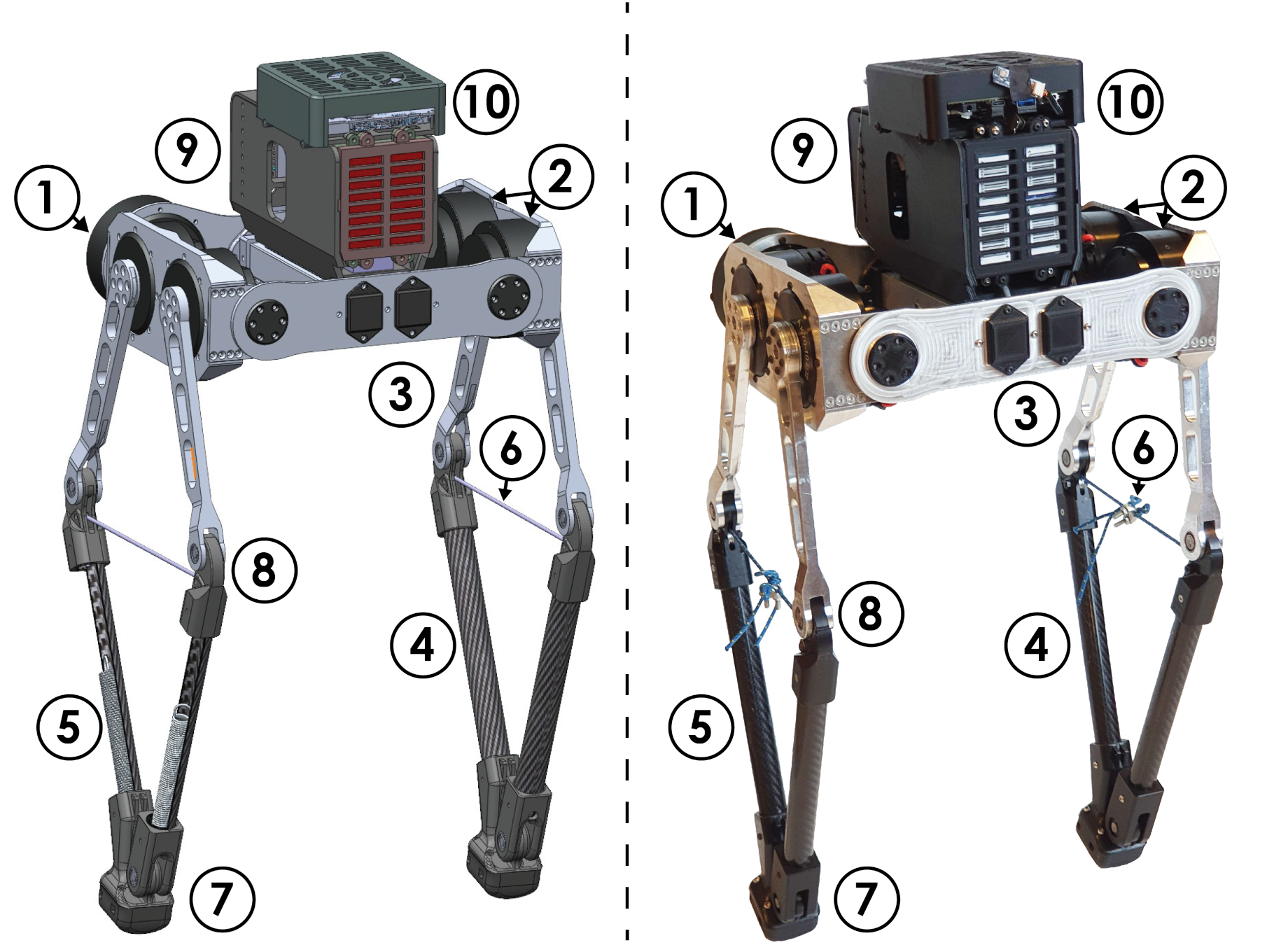}
    \vspace{-2ex} 
    \caption{Left: image of CAD model of bipedal robot realization, Right: the physical realized robot. Numbered elements: 1. Hip motor, 2. Motors for 5-bar linkage, 3. Hip link, 4. Calf link, 5. Embedded springs, 6. Paracord string, 7. Paw, 8. Passive knee joint, 9. Electronics unit, 10. Compute. Motors are numbered as follows, the left leg in the image contains Motor 1 (front), Motor 2 (center), and Motor 3 (back, hip motor). The right leg contains Motor 4 (front), Motor 5 (center), and Motor 6 (back, hip motor). }
    \label{fig:OLYMPUSCAD}
    \vspace{-5ex} 
\end{figure}

Towards a lightweight but mechanically robust design, the hip links were machined out of aluminum, and the calf links were made out of carbon fiber tubes. The springs are strategically placed inside the carbon fiber tubes as close to the paw end-effector as possible to lend their inertia close to the paw for future in-air stabilization movements. The springs are connected in a pulley system such that a high-strength paracord connects the ends of the two springs via a guide pulley on a ball bearing in the knee joint, connected to the other knee joint, as seen in Figure~\ref{fig:SimPoses} and \ref{fig:OLYMPUSCAD}. The joints connecting the aluminum and carbon tubes are machined Polyoxymethylene (POM), for robustness, and the paw is 3D printed out of TPU 95A. Located on top of the robot to avoid interference with hip actuation are the electronics, battery, and IMU needed for both the 6-motor bipedal realization, and also for a possible future 4-legged-12-motor version.

\section{SIMULATION STUDIES}\label{sec:sim}

Further jump simulations were performed in Simscape using the parameters of the realized bipedal robot to evaluate the vertical and forward jumping capacity for locomotion and overcoming significant obstacles using jumping. Although experimental studies are presented in the next section, it is noted that for these simulations, parameters such as joint friction were tuned to match the achieved real-life jumping results for the same motor torque and squat angle values. 

The primary objective of the simulations presented is to gain comprehensive insights into the dynamics of jumping in different gravity conditions and jump settings. We examine two distinct jump types, namely a) vertical jumps and b) forward jumps at a $30$-degree angle from vertical. The vertical and forward jumps are evaluated under varying motor power outputs and squat angles. The simulation under Earth gravity serves as the baseline, while the Mars scenario allows to explore how the reduced gravity impacts the jump. 

\begin{figure}[h!]
\centering
    \includegraphics[width=0.98\columnwidth]{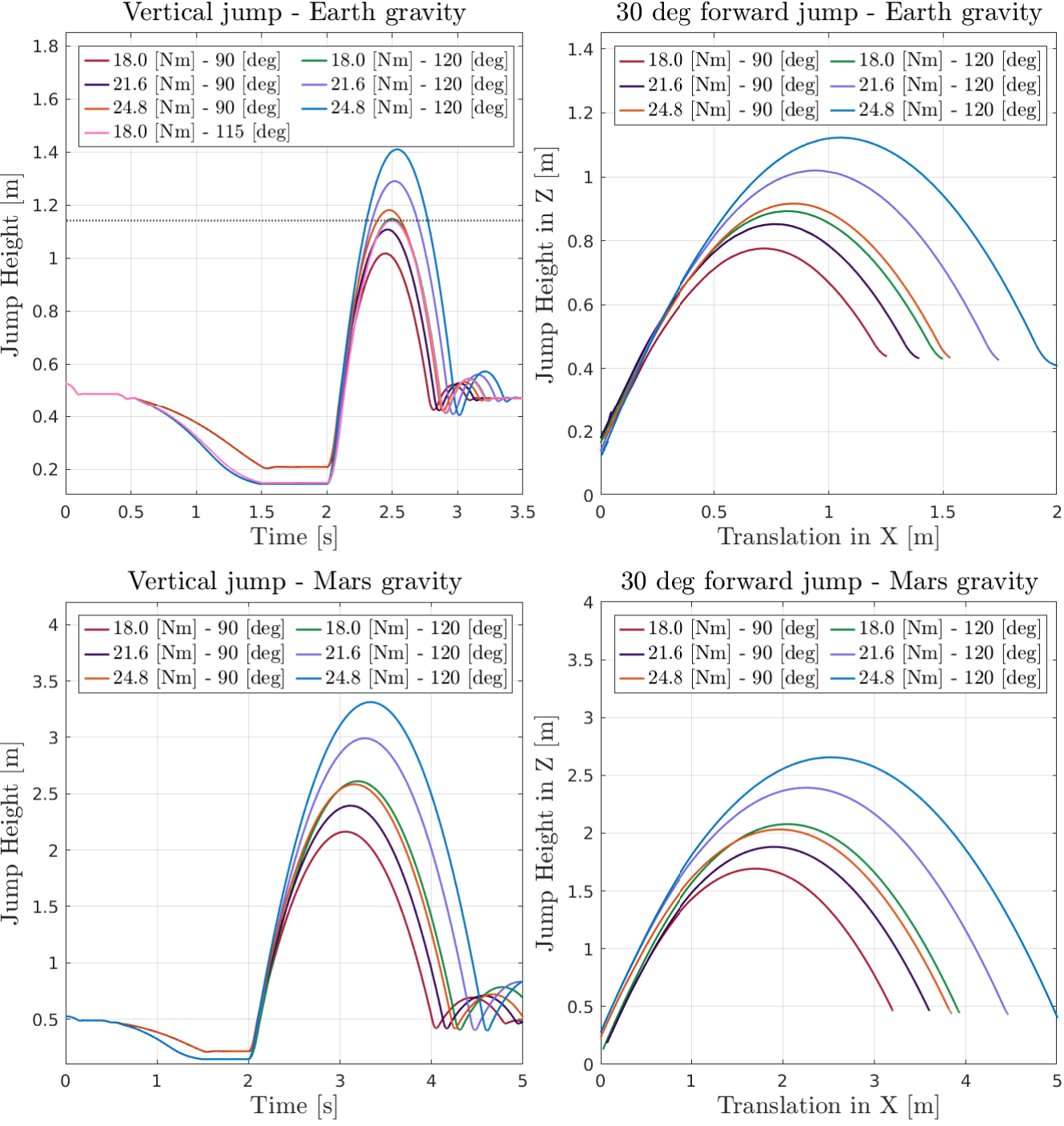}
        \vspace{-2ex} 
\caption{Simscape bipedal robot jump simulation for the chosen link lengths and spring stiffness.  Left side: vertical jumps. Right side: forward jumps at a 30-degree angle. The top plots are in Earth's gravity. The bottom plots are in Mars' gravity. Each plot encompasses jump simulations with six identical scenarios of torque and squat angle. On the upper left plot, the horizontal line highlights the $115\ \textrm{deg}, 18\ \textrm{Nm}$ result also tested experimentally.}
\label{fig:SimGravity_forward}
\vspace{-4.1ex}
\end{figure}

While in actual physical testing, significant safety margins are maintained to ensure the prototype robot's safety, the flexibility of simulation testing allows us to push these margins and explore closer to the robot's maximum jump performance. In Figure \ref{fig:SimGravity_forward}, jump heights and jump distances are illustrated for Earth and Mars gravity. Using the parameters as in the real robot, including spring coefficient of $435 \textrm{N/m} $, and setting the motor torque and the squat angle to maximum ($24.8\ \textrm{Nm},~120\ \textrm{deg}$) the maximum height in Earth gravity is  $1.41\ \textrm{m}$, while on Mars gravity this is increased to $3.31 \ \textrm{m}$. In Mars' gravity a maximum forward jump of $4.95\ \textrm{m}$ with an apogee height of $2.65\ \textrm{m}$ is reached.

It is noted that as described in Section~\ref{sec:bipedalrealization}, the realized robot utilizes a spring coefficient of $435\ \textrm{N/m} $. This is despite the fact that a spring value around $800 \ \textrm{N/m}$ would offer higher jumping, and represents a decision driven by the need to achieve a more damped and thus safer experimental jumping response. Nevertheless, a simulation was also conducted for a more performant spring rating (specifically $870\ \textrm{N/m}$) and also the maximum motor torque and squat angle values. In this scenario, the resulting jump altitude for the robot is determined to be $1.52\ \textrm{m}$ and a ground clearance of $1.06\ \textrm{m}$ for the paw while the leg is in $h_{\textrm{nominal}}$ in air. For Mars, the jump height is $3.63 \ \textrm{m}$ for the body and $3.17 \ \textrm{m}$ for the paw.

\section{EXPERIMENTAL STUDIES}\label{sec:experiments}
To validate and evaluate the functionality and capabilities of the proposed leg design and the complete bipedal system, several experiments were performed. A motion capture system established ground truth for tracking during tests. 

\vspace{-0.5ex}
\subsection{Leg Workspace}
\begin{figure}[h!]
    \centering
    \includegraphics[width = 0.99\columnwidth]{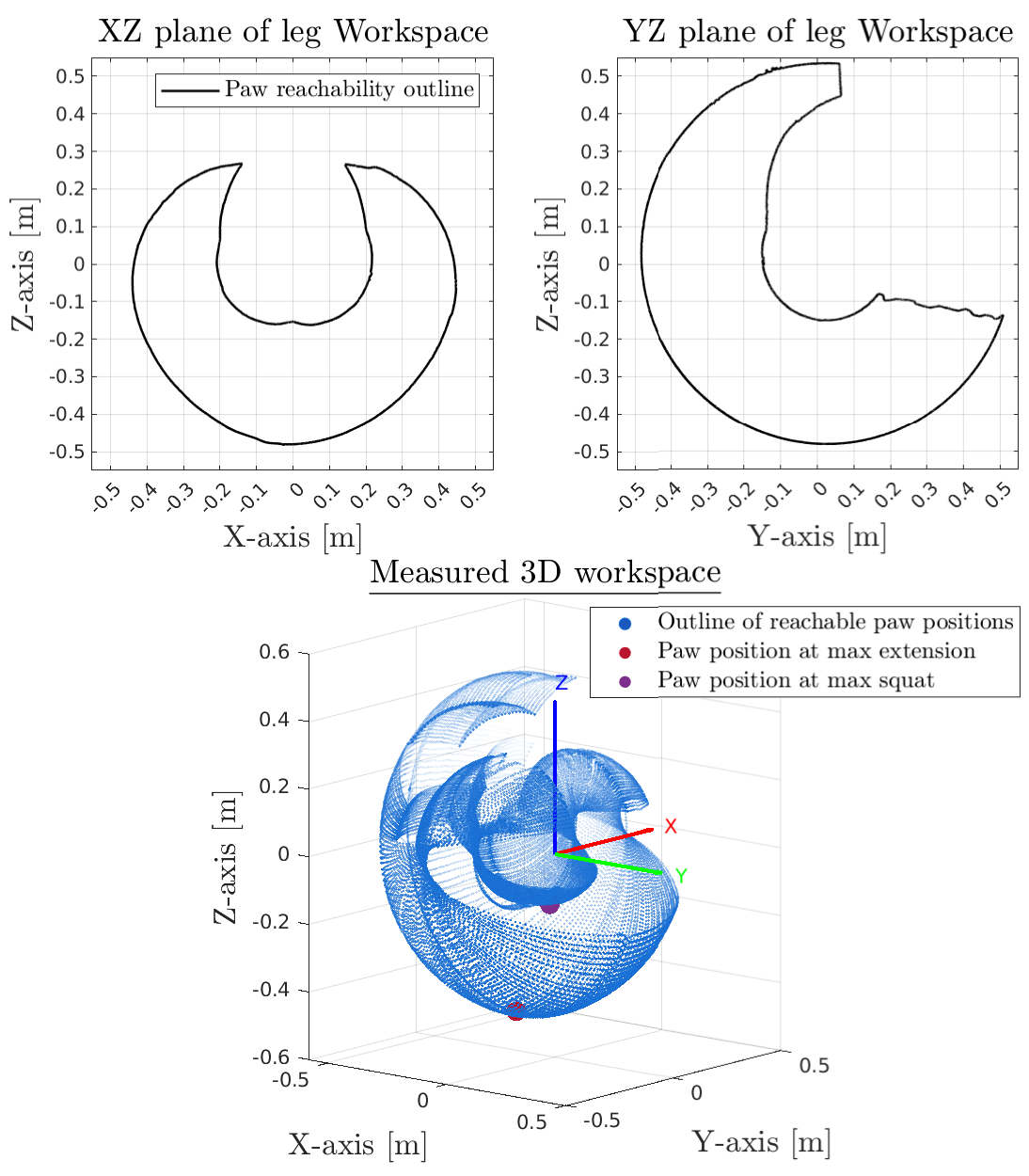}
    \vspace{-2ex}
    \caption{Outline of the experimentally verified limits of the workspace for one leg. The paw positions at max squat and at the end of the jump are marked. }
    \label{fig:workspace}
    \vspace{-3ex}
\end{figure}

The robot needs to have a workspace with three degrees of freedom per leg to enable fast and large dynamic maneuvers to correct angular velocity and attitude mid-air. To illustrate the limits of the workspace of each leg, the maximum angles achievable for each motor were set to Motor 1: $[-157,160]\ \textrm{deg}$, Motor 2: $[20, 337]\ \textrm{deg}$. Then, each leg performs a workspace movement outlining the area which the end effector can reach. Subsequently, actuating Motor 3 (hip actuator) in the range $[-70,180]\ \textrm{deg}$ with $2.5 \ \textrm{deg}$ increments, the $3$D workspace of the robot's leg is acquired. The resulting workspace, including both detailed $2\textrm{D}$ slices and the collective $3\textrm{D}$ visualization, can be seen in Figure~\ref{fig:workspace}.

\subsection{Full robot vertical jump}

Preliminarily, each individual leg, and subsequently the bipedal system were attached on a rail in order to tune the PID controllers of the motors. However, testing on rails introduces a set of limitations and impacts the fidelity of the results due to phenomena such as friction. Therefore, in order to evaluate the true jumping capacity of the designed leg and the bipedal system, vertical leaps were conducted free from supportive rails using only suspension with ropes for safety catching and assisting in balancing pre-jump. During these tests, the bipedal system as described in Section~\ref{sec:bipedalrealization} was employed and integrated all the computational, electronic, and mechanical systems, as well as the onboard battery, and thus also had its total mass. The robot uses both legs for a synchronized jump, showcasing its leaping prowess, control, and robust mechanical design. It is noted that for purposes of testing safety, the maximum motor torque values combined with the maximum squatting angles were not used but values close to the maxima were tested. Likewise, the coefficient of the integrated spring is not leading to true maximum jumping but allows good performance combined with a more damped response in both small and powerful leaps or walking. 

\begin{figure*}[h]
    \centering
    \includegraphics[width = \textwidth] {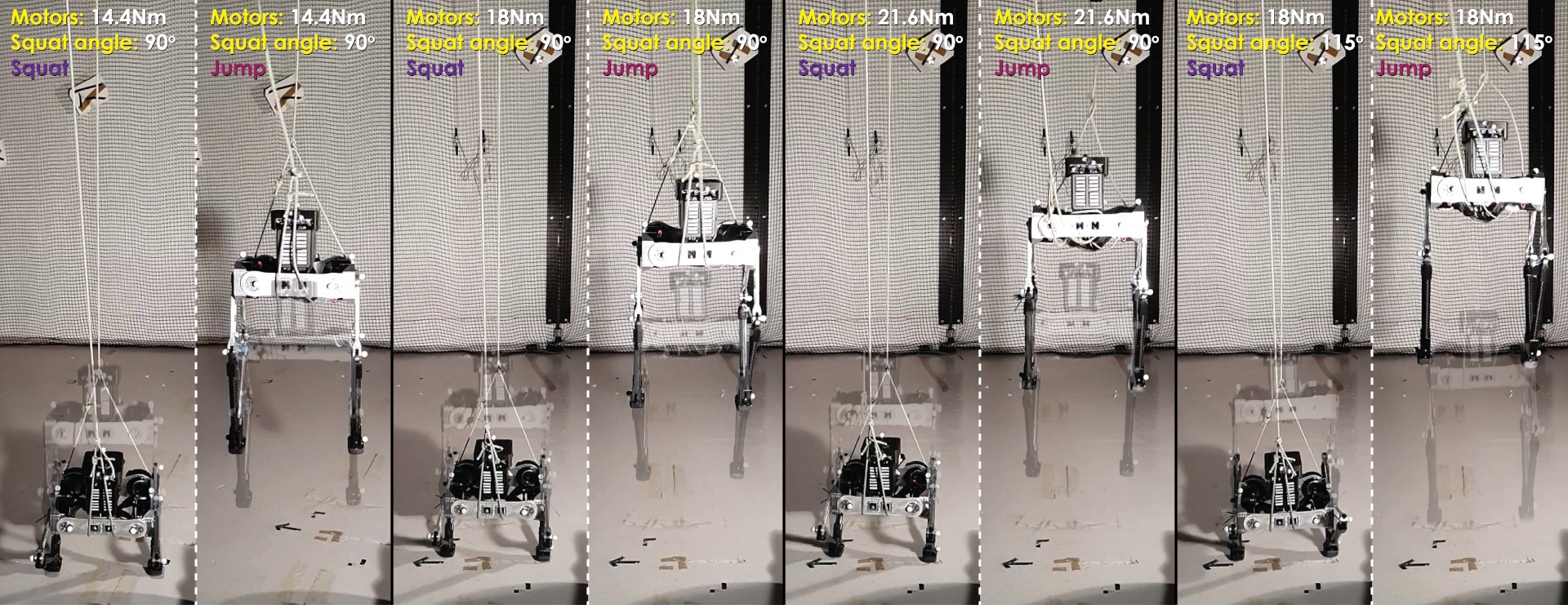}
    \vspace{-5ex}
    \caption{Image sequences depicting the squat and jump phase of $4$ experimental studies, namely (from left to right): 1) motors' torque at $14.4 \ \textrm{Nm}$ with squat angle of $90 \ \textrm{deg}$ jumping to $0.92 \ \textrm{m}$, 2) motors' torque at $18 \ \textrm{Nm}$ with squat angle of $90 \ \textrm{deg}$ jumping to $1.01 \ \textrm{m}$, 3) motors' torque at $21.6 \ \textrm{Nm}$ with squat angle of $90 \ \textrm{deg}$ jumping to $1.07 \ \textrm{m}$, and 4) motors' torque at $18 \ \textrm{Nm}$ with squat angle of $115 \ \textrm{deg}$ jumping to $1.141 \ \textrm{m}$. For the last three experiments, the relative camera pose is similar.}
    \label{fig:collective_jumps}
    \vspace{-3ex}
\end{figure*}

In total four experimental free jumping results (without rails) are presented and Figure~\ref{fig:collective_jumps} shows images from these tests with varying squat angle and torque saturation. The last experiment shown is the one for the maximum jump tested.

\begin{figure}[h]
    \centering
    \includegraphics[width = 0.99\columnwidth]{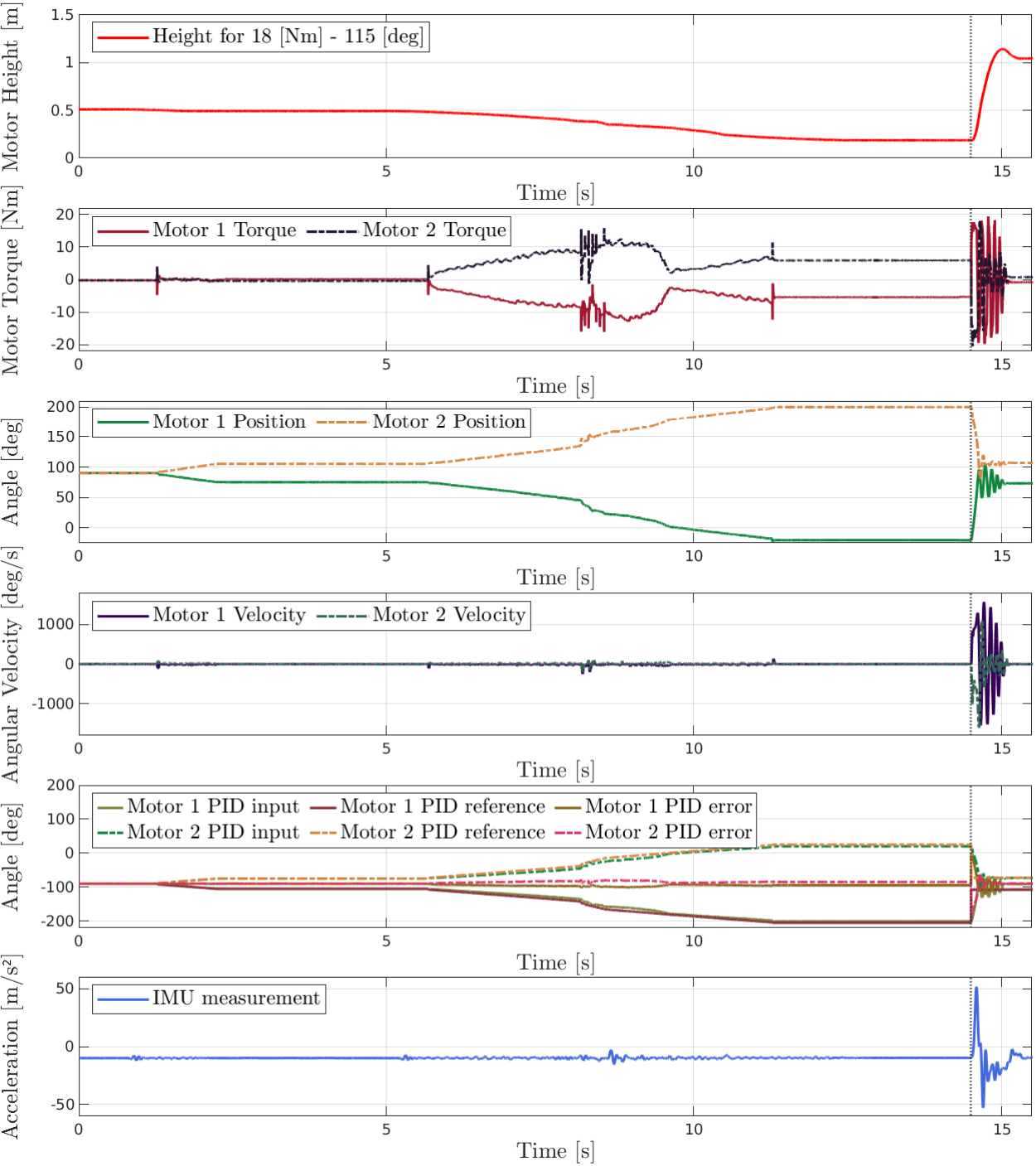}
    \vspace{-4.5ex}
    \caption{Jump test with motors' torque saturation set at $18\ \textrm{Nm}$ and a squat angle of $115\ \textrm{deg}$. The subplots illustrate the following during jump: 1. Jump height, 2. Motor Torque for the motors of one of the legs, 3. Associated motor angles, 4. Motor angular velocity, 5. Motor PID controllers response indicating good tracking at all moments other jump initiation. 6. IMU-measured acceleration passed through a lowpass filter with a time constant $10\ \textrm{ms}$. The dotted line at $14.5$\ \textrm{s} indicates the time of jump initiation. }
    \label{fig:maxexperimentaljump}
    \vspace{-2ex}
\end{figure}

In further detail, setting the motor torque saturation at $18\ \textrm{Nm}$ ($72.5\%$ of maximum) and the squat angle to $115\ \textrm{deg}$ ($96\%$ of maximum), the robot during this experimental test reached a jump height of $1.141 \ \textrm{m}$ measured from the robot's main body (as seen in Figure~\ref{fig:SimPoses}) and ground clearance for the paw of $0.7 \ \textrm{m}$. 
Figure~\ref{fig:maxexperimentaljump} details this result. In this figure, the motor states are for only motor one and motor two is shown in the subplots due to the behavior of motor 4 and 5 are equal but in opposite directions, and states for motor 3 and 6 are not significant to note as they only hold their position through the jump.

\begin{figure}[h]
    \centering
    \includegraphics[width = 0.99\columnwidth]{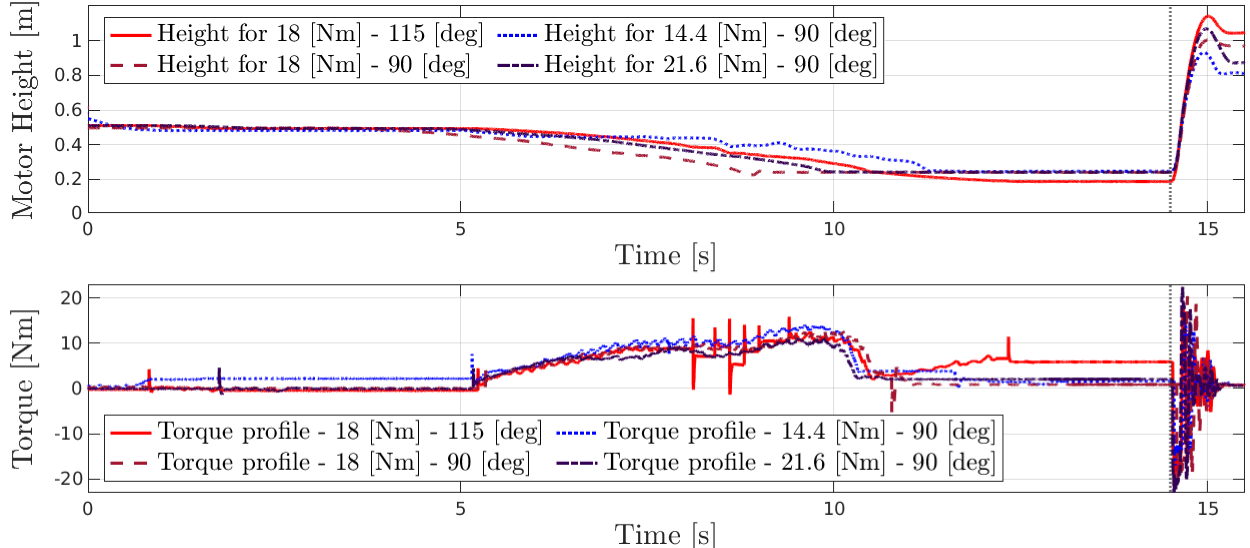}
    \vspace{-4.5ex}
    \caption{Jump responses of the four outlined experiments, alongside the torque response of Motor 2 of the robot during each of those tests. The dotted line at $14.5\ \textrm{s}$ indicates the time of jump initiation. }
    \label{fig:jumpState}
    \vspace{-2ex}
\end{figure}

Beyond this result, Figure~\ref{fig:jumpState} depicts the jumping maneuver trajectories for all four experiments illustrated in Figure~\ref{fig:collective_jumps}. In particular, the three first results were conducted with motor torque-and-squat angle settings at: $[14.4\ \textrm{Nm},90\ \textrm{deg}]$, $[18\ \textrm{Nm},90\ \textrm{deg}]$,  $[21.6\ \textrm{Nm},90\ \textrm{deg}]$, while the last one (at $[18\ \textrm{Nm},115\ \textrm{deg}]$) is the one further outlined before.

\section{CONCLUSIONS AND FUTURE WORK}\label{sec:conclusions}

This paper presented the design, simulation, and testing of a legged system optimized for jumping in low gravity. Employing a 5-bar linkage as a basis for leg concept, the investigation into link lengths resulted in a bipedal design optimized for Mars' gravity. The robot realization proved robust and effective at jumping and demonstrated a significant workspace which is necessary for future work on in-air stabilization and dynamic walking. With significant safety margins in spring stiffness, motor torque, and squat angle, the real robot achieved a maximum jump height of $ 1.141 \ \textrm{m}$. Simulations show that this can be extended in Earth gravity with lesser safety margins and reach a height of $3.63 \ \textrm{m}$ in Mars gravity, highlighting the effectiveness of using jumping to overcome significant obstacles in such low gravity environments. Future work includes further testing and development, developing optimal jump controllers, and possibly extending the design to a quadrupedal system.

\bibliographystyle{IEEEtran}

\bibliography{./BIB/ICRA2023.bib}

\end{document}